\documentclass{IOS-Book-Article}

\usepackage{mathptmx}
\usepackage{booktabs} 
\usepackage{soul}\setuldepth{article}
\usepackage[table]{xcolor}

\usepackage{caption}
\usepackage{subcaption}
\def\hb{\hbox to 11.5 cm{}}
\usepackage{multirow}

\usepackage{graphicx}
\usepackage{xcolor}
\usepackage{amssymb}

\makeatletter
\newcommand\footnoteref[1]{\protected@xdef\@thefnmark{\ref{#1}}\@footnotemark}
\makeatother

\begin{document}

\pagestyle{headings}
\def\thepage{}
\begin{frontmatter}             
\title{Harnessing GPT-3.5-turbo for Rhetorical Role Prediction in Legal Cases}

\markboth{}{September 2023\hb}

\author[A]{\fnms{Anas} \snm{belfathi}%
},
\author[A]{\fnms{Nicolas} \snm{hernandez}}
and
\author[A]{\fnms{Laura} \snm{monceaux}}

\address[A]{Nantes Université, École Centrale Nantes, CNRS, LS2N, UMR 6004, 
France}

\begin{abstract}

We propose a comprehensive study of one-stage elicitation techniques for querying a large pre-trained generative transformer (GPT-3.5-turbo) in the rhetorical role prediction task of legal cases.
This task is known as requiring textual context to be addressed. 
Our study explores 
strategies such as zero-few shots, task specification with definitions and clarification of annotation ambiguities, textual context and reasoning with general prompts and specific questions.   
We show that the number of examples, the definition of labels, the presentation of the (labelled) textual context and specific questions about this context have a positive influence on the performance of the model.
Given non-equivalent test set configurations, we observed that prompting with a few labelled examples from direct context can lead the model to a better performance than a supervised fined-tuned multi-class classifier based on the BERT encoder  (weighted F1 score of~$\approx$72\%). 
But there is still a gap to reach the performance of the best systems ~$\approx$86\%) in the LegalEval 2023 task which, on the other hand, require dedicated resources, architectures and training.

\end{abstract}

\end{frontmatter}
\markboth{September 2023\hb}{September 2023\hb}

\section{Introduction}

Large Language Models (LLMs) have proved effective for a variety of applications, but adapting them to a task or a specialized domain remains a major challenge. 
In recent years, prompting Generative LLMs 
has become a dominant paradigm as a first approach to solving various downstream tasks \cite{brown-2020-few-shot-learners}. 
However, few studies have focused on the task of rhetorical role prediction using generative approaches, in particular in legal cases \cite{savelka2023gpt4}.
Kalamkar et~al. showed that labelling sentences of legal cases with rhetorical roles, such as Facts, Arguments or Analysis, improve performance on the tasks of summarization and legal judgment prediction~\cite{kalamkar2022corpus}. 

This research paper focuses on evaluating the potential of generative pre-trained transformers (GPT), specifically OpenAI's GPT-3.5-turbo \cite{brown-2020-few-shot-learners,ouyang2022instructgpt}, to autonomously conduct rhetorical analysis on sentences extracted from legal cases.
Generative approaches in the legal domain are appealing because adapting a LLM by fine-tuning it requires annotated data, which is expensive to produce for each court in each country. 
In addition, predicting the rhetorical label of a sentence requires taking into account its textual context, and even the text as a whole. But the capacity of the state-of-the-art LLMs does not always allow them to take into account the entirety of a legal case~\cite{belfathi-2023-spe}. 
The state-of-the-art systems define the problem as a sequence labelling task~\cite{modi2023legaleval}. 
Because of the generative aspect of the GPT-3.5-turbo model and inspired by Savelka et~al.~\cite{savelka2023gpt4}, we define the problem as a multi-classification task through experimentation with several prompting strategies, such as Zero-Few shot prompting, Chain-of-Thought reasoning, specialized legal knowledge specification and textual context prompting.

We propose the following research questions: 
(RQ1) To what extent can GPT-3.5-turbo successfully perform labelling tasks using classical prompting techniques by giving \textit{zero, one or a few examples}?
(RQ2) To what extent does providing the \textit{label definitions} influence the efficiency and accuracy of GPT-3.5-turbo, and how does the model benefit from the inclusion of \textit{clarification of ambiguities between labels}?
(RQ3) What are the implications of utilizing \textit{textual context} in the label prediction of a sentence?
(RQ4) How does prompting the model to \textit{think step by step and to explain its reasoning} without specifying particular expectations affect its performance and response quality in comparison to asking \textit{precise questions about the textual context}?

\section{Related works}

\textit{Prompt engineering}, also known as \textit{in-context prompting}, refers to techniques aiming at steering the Generative LLM's behavior towards a particular outcome without updating the model's parameters \cite{huang-chang-2023-towards,qiao-2023-reasoning}. 
The most basic technique, called \textit{zero-shot prompting}, consists of feeding the model with a request and asking for completion. This technique can be enhanced by offering one or a few examples of input–output pairs
 in the prompt that guide the model to carry out the task; The technique is so called \textit{few-shot prompting}.
Brown et~al.~\cite{brown-2020-few-shot-learners} and Wei et~al.~\cite{wei-2022-finetuned-llm-are-zero-shot,wei-2022-emergent-abilities} demonstrated the ability of LLMs with more than 100 billion parameters (such as 175B GPT-3) to respond successfully to such requests for several tasks, with even better results when the models were fine-tuned to respond to instructions (such as 175B GPT-3.5-turbo).
By investigating GPT-3 on few-shot classification tasks, Zhao et~al.~\cite{zhao-2021-calibrate} demonstrated that the choice of the prompt format, the training examples, and the order of the examples can affect the accuracy of the results.
To select the examples, Liu et~al.~\cite{liu-2022-good-examples} recommend to retrieve examples that are semantically similar to the test example and Diao et~al.~\cite{diao-2023-active} supplement by showing that  examples with high disagreement or entropy (from a set of candidate examples) are among the most important and useful.
Lu et al.~\cite{lu-2022-ordered-prompts} observed that generative models (like 175B GPT-3) 
are sensitive to the examples ordering whatever the model size or the number of examples. 
In their approach, called In-Context Instruction Learning (ICIL), Ye et al.~\cite{ye-2023-incontext-instruction} showed that providing a fixed prompt with multiple cross-task demonstrations\footnote{Where each demonstration is a concatenation of an instruction, input, and output instance of a task.} as context of a third-party task query enhance the model performance on several tasks.
The authors suggest that effectiveness 
comes from 1) selecting classification tasks that include explicit answer choice in the instruction and 2) retrieving demonstrations that are similar to the target task.
Recent works have shown that explaining the reasoning or \textit{Chain-of-Thought} (CoT), required to solve a task, increases the performance of generative models in solving the task \cite{wei2023chainofthought}.
Reasoning can be seen as decomposing a problem into a sequence of sub-problems either iteratively or recursively \cite{mialon-2023-augmented}. 
Surprisingly, simply encouraging the model to reason (by adding "Let's think step by step" before an answer) 
can also improve the generation \cite{kojima-2022-zero-shot-reasoners}. 
Ye and Durrett~\cite{ye2022explanationsinreasoning} have shown that GPT-3.5 benefits substantially from prompting with explanations for reasoning over text (question answering and entailment).
Fu et~al.~\cite{fu2023complexitybased} have shown that prompts with higher reasoning complexity (i.e. chains with more reasoning steps) achieve better performance than simple prompts on math word reasoning tasks.

\paragraph{Generative LLM for legal tasks.}
Measuring the influence of generative models on legal tasks has become one of the main concerns of NLP researchers working in this field \cite{savelka2023unlocking,savelka2023gpt4,yu2023promprengineeringlegalreasoning}.  
On a legal entailment task (question answering task based on a legal article of  a few sentences), Yu et~al.~\cite{yu2023promprengineeringlegalreasoning} showed that giving the article in the prompt and asking a GPT-3 model to analyse it according to a given rhetorical schema (corresponding to a legal reasoning approach) improved performance compared to few-shot examples techniques or a zero-shot CoT \cite{kojima-2022-zero-shot-reasoners} strategy.
Savelka et al. \cite{savelka2023gpt4} questioned the use of GPT-4 for multi-class sentence classification tasks on US court opinions.
They experimented with prompts containing annotation guidelines originally designed for human annotators, with clarifications of ambiguities between labels and with requests for explanations. 
They did not measure the contribution of the annotation guidelines but they observed that 
the model performance is comparable to that of the best-performing law student annotators. 
They showed that disambiguation of labels enhance the performance. 
Eventually they found that asking the model to explain its choice of label reduces performance. 

\paragraph{The rhetorical role prediction task}
The SemEval 2023 LegalEval shared task \cite{modi2023legaleval} provides a good insight of the dominant approach in addressing the rhetorical role prediction task.
Most of the participants defined the problem as a sequence sentence classification task
and adopted a system architecture based on the Hierarchical Sequential labelling Network (HSLN) \cite{kalamkar2022corpus,brack2021sequential}, denoted as \textit{SciBERT-HSLN}. 
The best system \cite{huo-2023-antcontenttech}, denoted \textit{AntContentTech}, equipped \textit{SciBERT-HSLN} with domain-adaptive pretraining, data augmentation strategies, as well as auxiliary-task learning techniques.
On the LegalEval test dataset, \textit{SciBERT-HSLN} obtained a weighted F1 score of 0.79 while \textit{AntContentTech} obtained a score of 0.8593. For an indicative comparison, \cite{belfathi-2023-spe}\footnote{Since the authors did not have access to LegalEval test dataset, training, validation and test were performed on 80/10/10 splits from the concatenation of the train and dev LegalEval datasets.} reported a score of 0.65 with a simple BERT fine-tuned for single sentence classification (hereinafter denoted as \textit{BERT}), and between 0.75-0.77 with architecture adaptations to take into account the local context of the sentence to label (denoted as \textit{BERT+local\_context}).

\section{Exploring Various GPT Prompting Strategies}

Our prompts are inspired from \cite{savelka2023gpt4}. The template is made of four parts:  \texttt{PREAMBULE}, \texttt{EXAMPLES-or-CoT}, \texttt{INPUT} and \texttt{REQUEST}. 
The \texttt{PREAMBLE} is common to all prompts and takes 211 tokens (See Figure~\ref{prompt-parts}~(a)). It sets the persona and the domain and the task definition of the forthcoming \texttt{REQUEST}. 
The \texttt{EXAMPLES-or-CoT} is the more volatile part of the prompt. It will be presented in detail in the following sections.
The \texttt{INPUT} part has a simple form: "\texttt{SENTENCE: $\backslash n$```\{sentence\}'''}". 
And so has the \texttt{REQUEST} part:  "\texttt{EXPECTED OUTPUT FORMAT: $\backslash n$Label: $<$label$>$}".
In practice, 
this part may be more specified depending on the experiment.
The presentation of the task, the label definitions and the bootstrapping examples come from the LegalEval 2023 task~\cite{kalamkar2022corpus}. 

\begin{figure}[th]
  \tiny
  \begin{tabular}{c|l|}

\toprule 
\multirow{13}{*}{(a)} & \multirow{1}{11cm}{
{\fontfamily{cmss}\selectfont
You are a specialized system focused on semantic annotation of court opinion.\\ 
 $\backslash$n RHETORICAL ROLE:\\
Rhetorical roles in legal writing refer to the distinct functions or purposes that different parts of a document, such as a legal opinion, serve in conveying information, persuading the reader, and constructing a coherent argument. These roles encompass various elements like factual background, legal principles, arguments, counter arguments, and conclusions, each contributing to the document's overall persuasive and informative structure. \\
 $\backslash$n labelling TASK:\\
Please label each sentence in the document with one of the following predefined rhetorical roles:  'Preamble', 'Facts', 'Ruling by Lower Court', 'Issues', 'Argument by Petitioner', 'Argument by Respondent', 'Analysis', 'Statute', 'Precedent Relied', 'Precedent Not Relied', 'Ratio of the decision', 'Ruling by Present Court', 'NONE'.\\
Assign the role that best describes the purpose or function of each sentence in the context of the legal opinion.
}}
\\
& \\
& \\
& \\
& \\
& \\
& \\
& \\
& \\
& \\
& \\
\midrule

\multirow{6}{*}{(b)}& \multirow{1}{11cm}{
{\fontfamily{cmss}\selectfont
EXAMPLES:\\
SENTENCE: IN THE COURT OF THE IV ADDL SESSIONS JUDGE, CHENNAI. Dated this the 10th day of September 2023.\\
LABEL: Preamble\\
SENTENCE: SUPREME COURT OF INDIA. Dated this the 5th day of June 2022. This judgment pertains to the case of John Doe versus Jane Smith.\\
LABEL: Preamble$\backslash$ n
$[...]$
}}
\\

& \\
& \\
& \\
& \\
& \\
& \\
\midrule 
\multirow{8}{*}{(c)}& \multirow{1}{11cm}{
{\fontfamily{cmss}\selectfont
ANNOTATION GUIDELINES:\\
  - 'Preamble':A typical judgement would start with the court name, the details of parties, lawyers and judges' names, Headnotes. This section typically would end with a keyword like (JUDGEMENT or ORDER etc.). Some supreme court cases also have HEADNOTES, ACTS section. They are also part of Preamble.\\
  - 'Issues': Some judgements mention the key points on which the verdict needs to be delivered. Such Legal Questions Framed by the Court are ISSUES. $\backslash$ n
E.g. “he point emerge for determination is as follow:- (i) Whether on 06.08.2017 the accused persons in furtherance of their common intention intentionally caused the death of the deceased by assaulting him by means of axe ?”$\backslash$ n
$[...]$
}}

\\
& \\
& \\
& \\
& \\
& \\

& \\
& \\
\midrule

\multirow{9}{*}{(d)} & \multirow{1}{11cm}{
{\fontfamily{cmss}\selectfont 
ANNOTATORS QUALITY ASSESSMENT:\\
It is important to note that during the annotation process, certain patterns emerged in annotators' assessments:\\
$*$ High Agreement: Amongst annotators, high agreement was observed for 'Preamble', 'Ruling by Present Court', 'NONE', and 'Issues'.\\
$*$ Medium Agreements: For 'Facts', 'Ruling by Lower Court', 'Analysis', 'Precedent Relied', and 'Argument by Petitioner' and 'Argument by Respondent', medium agreements were noted.$\backslash$ n
$[...]$
}}
\\

& \\
& \\
& \\
& \\
& \\
\midrule

\multirow{8}{*}{(e)}& \multirow{1}{11cm}{
{\fontfamily{cmss}\selectfont 
CONTEXT SENTENCES:\\
SENTENCE: ```It entered into transactions in the nature of forward transactions with parties at Bhatinda (in the Patiala State outside the taxable territories of British India) in which it suffered losses.``` \\
LABEL: Preamble\\ 
SENTENCE: ```The assessee claimed deduction of these losses in the computation of its income.```\\
LABEL: Preamble $\backslash$ n
$[...]$\\
}}
\\
& \\
& \\
& \\
& \\
& \\
\midrule
\multirow{2}{*}{(f)}& \multirow{1}{11cm}{
{\fontfamily{cmss}\selectfont 
 EXPECTED OUTPUT FORMAT(Give your response in a json format. Stick to less than 30 words):\\
  \{\{$\backslash$n  "Let’s think step by step": \textless reasoning why particular label should be assigned \textgreater$\backslash$n\\   
  "Label": \textless label \textgreater $\backslash$n 
  \}\}
}}
 \\

& \\
& \\
\midrule
\multirow{7}{*}{(g)}& \multirow{1}{11cm}{
{\fontfamily{cmss}\selectfont
EXPECTED OUTPUT FORMAT(Give your response in a json format. Stick to less than 30 words):\\
  \{\{$\backslash$n  "Label": \textless label \textgreater, $\backslash$n \\
  "Relative position": "\textless return the relative position corresponding to the most sentence presented in the context (NEXT SENTENCES) that impact more on the decision\textgreater", $\backslash$n \\
  "Sentence": "\textless return the full text of the impacted sentence corresponding to the relative position\textgreater", $\backslash$n \\
  "Terms": "\textless List up to 5 words from the context and the predicted sentence that significantly influence the decision, in array format\textgreater"$\backslash$n 
  \}\}
  ""”
  }}

\\
& \\
& \\

& \\
& \\
& \\
& \\\bottomrule
  \end{tabular}
  \caption{Various possible parts of the prompts: PREAMBLE (a);
  \texttt{EXAMPLES-or-CoT}  
  with Few-Shot Prompting (b),  
  with Label definitions 
  (c), 
  with Clarification of label ambiguities
  (d) in extension of (c), 
  with labelled textual context (e);
  \texttt{REQUEST} to encourage the model to reason (f), with specific questions (g).
  }
\label{prompt-parts}
\end{figure}

\subsection{Zero-Few Shot Prompting (RQ1)}

Our first experiment was to assess the proficiency of GPT in performing our task using zero-few-shot prompting.
For \textit{zero-shot} prompting, we left 
the \texttt{EXAMPLES-or-CoT} part empty. 
For one and more shot prompting,
we left the problem of selecting significant examples for future work. Instead we asked GPT-3.5 to generate examples by taking inspiration from  the  examples given in the explanatory Figure about the Rhetorical Roles of~\cite{kalamkar2022corpus}.
We assumed that they were representative and that the model would better understand something that it had generated itself.
The prompt we used for generate the examples was: \texttt{Given these examples of each Rhetorical Roles label, generate four representative sentences for each label}. 
We limited the generation to four sets of examples due to the input length limits of the model (one shot was about 850 tokens).
Figure~\ref{prompt-parts}~(b) shows an illustration of the use of examples generated for a two-shot prompting.


\subsection{Label Definitions and Clarification Between Labels (RQ2)}
\label{labeldefinitionssection}

In this experiment, we sought to explore the impact of  providing label definitions,
possibly supplemented by the clarification from the annotator errors (denoted as \textit{definition+clarifications}). 
So the \texttt{EXAMPLE-or-CoT} part of the prompt was first fed with the label definitions provided by Kalamkar et~al.~\cite{kalamkar2022corpus} in table of an appendix of the online resources\footnote{\url{https://github.com/Legal-NLP-EkStep/rhetorical-role-baseline}} (See Figure~\ref{prompt-parts}~(c)). Subsequently, to address the consequences of introducing clarifications about the annotator errors and ambiguities between labels to the GPT model, we extended the label definitions with  
the content of the "Annotation Quality Assessment" section of \cite{kalamkar2022corpus} (See Figure~\ref{prompt-parts}~(d)). To ensure alignment with the narrative of our prompt message, we made some minor modifications, including the removal of references and the organization of the paragraph into distinct points, each addressing separate ambiguities. 
In order to have a complete view, we also have combined the definitions with four-shot examples.


\subsection{Textual Context Enrichment (RQ3)}
\label{labelledtextualcontextsection}

The main objective of this experiment was to examine the impact of presenting the textual context of an input sentence to the model.
The coherence of a text is expressed by the fact that consecutive sentences are linked by thematic and rhetorical relationships. The underlying idea is to get the model to exploit this information.
We also discuss the fact of providing the labels of the sentences in this context. 
Indeed the experiment can be seen as a variant of few-shot learning where examples are selected for certain reasons (their belonging to the textual context of the target sentence), possibly coming without labels.
Based on Belfathi et~al.~\cite{belfathi-2023-spe}, we studied both
the direction
and the size of the context window to consider. 
In practice, we experimented with adding 2 or 8 preceding or following sentences 
in the \texttt{EXAMPLE-or-CoT} part of the prompt (See Figure~\ref{prompt-parts}~(e)).

\subsection{Encouraging General or Specific Reasoning (RQ4)}

The aim of this experiment was to observe the behaviour of the model with general reasoning questions 
compared with specific questions. 
General questions were implemented 
by adding the expression "Let's think step by step"~\cite{kojima-2022-zero-shot-reasoners} into the \texttt{REQUEST} part of the prompt and by specifying we were expecting an explanation about the choice of a label by the model (See~Figure~\ref{prompt-parts}~(f))\footnote{\label{note1}
We instructed the model to provide the results in a JSON format to facilitate the evaluation process, and limited the token generation to 30 words to manage costs associated with the API usage.}. 
To probe the model's reasoning capacity with specific questions we targeted questions about the textual context.
Based on the configuration that obtained the best results in the experiment described in Section~\ref{labelledtextualcontextsection} (i.e. inserting the following 8 labelled sentences),
we asked the model about the relative position of the most influential sentence within the context and the relevant legal terms that impact the decision (See Figure~\ref{prompt-parts}~(g))\footnoteref{note1}.

\section{Experimental setting}

\paragraph{Data}

\begin{table}[ht]
\caption{Statistical Distribution: Percentage for Each Rhetorical Role in the LegalEval Corpus and Their Segments Used in Our Experimental Subset (\%)}
  \label{tab:cor-stat}
\setlength\tabcolsep{4pt} 
  \resizebox{\textwidth}{!}{%
  \begin{tabular}{lccccccccccccc} 
\hline
           \textbf{Dataset}         & \multicolumn{1}{l}{\textbf{Analysis}} & \multicolumn{1}{l}{\textbf{FACTS}} & \multicolumn{1}{l}{\textbf{PRMBL}} 
           & \multicolumn{1}{l}{\textbf{NONE}} & \multicolumn{1}{l}{\textbf{PRE-R.}} 
           & \multicolumn{1}{l}{\textbf{ARG-PET}} & \multicolumn{1}{l}{\textbf{RPC}} & \multicolumn{1}{l}{\textbf{RLC}} & \multicolumn{1}{l}{\textbf{RATIO}} & \multicolumn{1}{l}{\textbf{ARG-RES}} & \multicolumn{1}{l}{\textbf{STA}} & \multicolumn{1}{l}{\textbf{ISSUE}} & \multicolumn{1}{l}{\textbf{PRE-NOT-R.}} 
           \\ \hline
LegalEval    & 36.65                                 & 19.84                              & 14.67                                  & 5.06                              & 4.93                                    & 4.34                                 & 3.67                             & 2.72                             & 2.33                               & 2.30                                 & 1.59                             & 1.30                               & 0.53                                        \\
Experimental   & 30.33                                 & 18.34                              & 18.43                                  & 6.90                              & 7.81                                    & 1.63                                 & 3.63                             & 6.63                             & 1.72                               & 1.18                                 & 1.36                             & 1.72                               & 0.27                                        \\ \hline
\end{tabular}%
}
\end{table}

We utilized the data provided by Sub-task A, "Rhetorical Roles Prediction," of the SemEval 2023 Task 6, "LegalEval - Understanding Legal Texts" challenge~\cite{kalamkar2022corpus}\footnote{\url{https://sites.google.com/view/legaleval}}. This dataset consists of Indian legal data extracted from court judgments, featuring 13 distinct rhetorical roles (RRs). 
and averaging 117.31 sentences per document.
To prepare for our experiments, we randomly selected 10 documents (1,101 sentences) from the validation data to manage costs associated with using the GPT API.
We observed that there was less variation between the original LegalEval dataset and the segments chosen for our experimentation (See~Table~\ref{tab:cor-stat}).

\paragraph{Model parameters}

We experimented the \texttt{gpt-3.5-turbo} model (\texttt{-0613} snapshot from June 13th 2023) which extends \texttt{text-davinci-003} (175B GPT-3 LLM trained on code-completion tasks and fine-tuned on natural language instruction tasks) with optimization for chat\footnote{See~\url{https://platform.openai.com/docs/model-index-for-researchers} and \url{https://platform.openai.com/docs/models} for more details.}.
Its maximum input length is 4,096 tokens.  
To ensure that its completion was deterministic,
we set the temperature for all experiments to 0. Other parameters were set to their default values (Top P=1, Frequency penalty = 0, Presence penalty = 0). The cost of all experiments was 68 euros. 
\paragraph{Measures}

The performance of the NLP models for the rhetorical roles task is assessed using Weighted-Precision ($_wP$),  Weighted-Recall ($_wR$), Accuracy ($A$), Weighted-F1 ($_w{F1}$) and Macro F1 ($_M{F1}$) scores based on the hidden test set. The weighted F1 score considers both precision and recall, and it is calculated by taking into account the class-wise F1 scores weighted by the number of samples in each class. 
\section{Results and discussion}

\begin{table}[ht]
  \caption{Performance of Prompting Strategies Ordered by Research Question (RQ). Reported results for \textit{BERT} and \textit{BERT+local\_context} \cite{belfathi-2023-spe}, \textit{SciBERT-HSLN} \cite{kalamkar2022corpus}, and \textit{AntContentTech} \cite{huo-2023-antcontenttech} are given for information. All were trained on the same dataset source (LegalEval 2023) but with various splits and amounts of data.}
  \label{tab:mod-perf}
  \centering
  \setlength\tabcolsep{4pt}
  \resizebox{.5\textwidth}{!}{
  \begin{tabular}{clccccc} 
    \toprule
    \textbf{RQ} & \textbf{Model} & \textbf{$_wP$} & \textbf{$_wR$} & \textbf{$A$} & \textbf{$_w{F1}$} & \textbf{$_M{F1}$} \\
    \midrule
    \multirow{4}{*}{\textbf{1}} & zero-shot example & 0.42 & 0.34 & 0.34 & 0.33 & 0.29 \\
    & one-shot example & 0.45 & 0.33 & 0.33 & 0.33 & 0.30 \\
    & 2-shot example  & 0.45 & 0.34 & 0.34 & 0.36 & 0.29 \\
    & 4-shot example  & 0.46 & 0.35 & 0.35 & 0.37 & 0.28 \\
    \bottomrule
    \multirow{3}{*}{\textbf{2}} & definition & 0.46 & 0.42 & 0.42 & 0.42 & 0.33 \\
    & definition+clarification & 0.46 & 0.41 & 0.41 & 0.41 & 0.32 \\
    & definition+examples & 0.49 & 0.41 & 0.41 & 0.42 & 0.33 \\
    \bottomrule
    \multirow{8}{*}{\textbf{3}} & context-2 & 0.45 & 0.39 & 0.39 & 0.39 & 0.32 \\
    & context-8 & 0.45 & 0.36 & 0.37 & 0.36 & 0.29 \\
    & context+2 & 0.46 & 0.43 & 0.43 & 0.42 & 0.36 \\
    & context+8 & 0.43 & 0.39 & 0.39 & 0.38 & 0.31 \\
    & labelled\_context-2 & 0.66 & 0.63 & 0.63 & 0.63 & 0.50 \\
    & labelled\_context-8 & 0.71 & 0.68 & 0.68 & 0.68 & 0.50 \\
    & labelled\_context+2 & 0.69 & 0.66 & 0.66 & 0.66 & 0.51 \\
    & labelled\_context+8 & 0.72 & 0.70 & 0.70 & 0.70 & 0.53 \\
    \bottomrule
    \multirow{2}{*}{\textbf{4}} & zero-shot-cot  & 0.46 & 0.29 & 0.29 & 0.31 & 0.27 \\   
    & cot-by-queries & \textbf{0.77} & \textbf{0.71} & \textbf{0.71} & \textbf{0.72} & \textbf{0.61}\\
    \bottomrule
    &  \textit{BERT} \cite{belfathi-2023-spe} &  &  & & {0.65} & \\
    &  \textit{BERT+local\_context} \cite{belfathi-2023-spe}&  &  & & {0.75-0.77} & \\
    &  \textit{SciBERT-HSLN} \cite{kalamkar2022corpus} &  &  & & {0.79} & \\
    &  \textit{AntContentTech} \cite{huo-2023-antcontenttech} &  &  & & {0.8593} & \\
    \bottomrule
  \end{tabular}}
\end{table}

\begin{table}[ht]
  \centering
  \caption{Performance Measurement (F1 Score) of Models for Each Rhetorical Role Across All Experimentation Prompts. The blue cells signify the highest performance for each label.}
  \label{tab:perf-per-labels}
  \setlength\tabcolsep{4pt} 
  \resizebox{\textwidth}{!}{%
  \begin{tabular}{lccccccccccccc} 
    \toprule
    & \textbf{Analysis} & \textbf{ARG-PET} & \textbf{ARG-RES} & \textbf{FACTS} & \textbf{ISSUE} & \textbf{NONE} & \textbf{PRMBL} 
    & \textbf{PRE-NOT} 
    & \textbf{PRE\_R.} 
    & \textbf{RATIO} & \textbf{RLC} & \textbf{RPC} & \textbf{STA} \\
    \midrule
    zero-shot example                                    & 0.36                         & 0.21                        & 0.29                        & 0.46                      & 0.44                      & 0.27                     & 0.27                          & 0.29                                & 0.22                            & 0.00                      & 0.21                    & 0.50                    & 0.26                    \\
one-shot example                                     & 0.32                         & 0.12                        & 0.32                        & 0.48                      & 0.46                      & 0.27                     & 0.26                          & 0.31                                & 0.32                            & 0.00                      & 0.26                    & 0.51                    & 0.23                    \\
2-shot example                                     & 0.36                         & 0.20                        & 0.24                        & 0.45                      & 0.45                      & 0.22                     & 0.40                          & 0.22                                & 0.29                            & 0.00                      & 0.26                    & 0.47                    & 0.21                    \\
4-shot example                                     & 0.40                         & 0.17                        & 0.20                        & 0.47                     & 0.45                      & 0.25                     & 0.38                          & 0.12                                & 0.29                            & 0.00                      & 0.25                    & 0.48                    & 0.23                    \\

definition                                & 0.45                         & 0.26                        & 0.26                        & 0.54                      & 0.49                      & 0.13                     & 0.50                          & 0.20                                & 0.22                            & 0.09                      & 0.30                    & 0.58                    & 0.22                    \\
definition+clarification                                & 0.47                         & 0.21                        & 0.15                        & 0.55                      & 0.44                      & 0.17                     & 0.41                          & 0.25                                & 0.31                            & 0.08                      & 0.27                    & \cellcolor{blue!25}0.62                    & 0.26                    \\
definition+examples                                & 0.46                         & 0.23                        & 0.22                        & 0.53                      & 0.49                      & 0.23                     & 0.46                          & 0.17                                & 0.33                            & 0.04                      & 0.28                    & \cellcolor{blue!25}0.62                    & 0.20                    \\
context-2                      & 0.43                         & 0.26                        & 0.31                        & 0.55                      & 0.48                      & 0.22                     & 0.36                          & 0.24                                & 0.33                            & 0.07                      & 0.20                    & 0.46                    & 0.24                    \\
context-8                      & 0.38                         & 0.10                        & 0.28                        & 0.52                      & 0.51                      & 0.24                     & 0.36                          & 0.10                                & 0.29                            & 0.11                      & 0.23                    & 0.42                    & 0.25                    \\
context+2                      & 0.47                         & 0.22                        & 0.25                        & 0.55                      & 0.49                      & 0.24                     & 0.44                          & 0.44                                & 0.30                            & 0.09                      & 0.25                    & 0.53                    & 0.34                    \\
context+8                      & 0.43                         & 0.19                        & 0.19                        & 0.52                      & 0.41                      & 0.23                     & 0.39                          & 0.36                                & 0.27                            & 0.09                      & 0.19                    & 0.49                    & 0.30                    \\
labelled\_context-2                       & 0.67                         & 0.32                        & \cellcolor{blue!25}0.70                        & 0.74                      & 0.47                      & \cellcolor{blue!25}0.65                     & 0.71                          & 0.18                                & 0.50                            & 0.10                      & 0.37                    & 0.61                    & \cellcolor{blue!25}0.48                    \\
labelled\_context-8                      & 0.74                         & 0.34                        & 0.67                        & 0.79                      & 0.49                      & 0.54                     & 0.76                          & 0.00                                & 0.65                            & 0.00                      & 0.55                    & 0.54                    & 0.45                    \\
labelled\_context+2                       & 0.71                         & 0.30                        & 0.61                        & \cellcolor{blue!25}0.80                      & 0.49                      & 0.55                     & 0.77                          & 0.33                                & 0.58                            & 0.10                      & 0.45                    & 0.53                    & 0.38                    \\
labelled\_context+8                       & 0.78                         & 0.27                        & 0.44                        & 0.79                      & 0.59                      & 0.58                     & 0.78                          & 0.50                                & 0.64                            & 0.09                      & 0.58                    & 0.50                    & 0.38                    \\
cot-by-queries & \cellcolor{blue!25}0.79                         & \cellcolor{blue!25}0.36                        & 0.55                        & 0.75                      & \cellcolor{blue!25}0.61                      & 0.57                     & \cellcolor{blue!25}0.84                          & \cellcolor{blue!25}0.86                                & \cellcolor{blue!25}0.75                            & \cellcolor{blue!25}0.35                      & \cellcolor{blue!25}0.60                    & 0.46                    & 0.42                    \\
    \bottomrule
  \end{tabular}%
  } 
\end{table}

\subsection{Zero-Few Shot Prompting (RQ1)}

In this experiment, we examined GPT-3.5's efficiency in rhetorical role prediction within the legal domain utilizing \textit{zero} to {4-shot} prompting (See zero- and\textit{$[$1-4$]$-shot} examples in Table~\ref{tab:mod-perf}). 
The low Macro-F1 score indicates that the Zero-Few prompts encountered challenges in label recognition, often leading to confusion between different rhetorical role labels.
We can see that by increasing the number of examples, the Weighted-F1 score increases, but on the other hand the Macro-F1 score slightly decreases. 
As confirmed by Table~\ref{tab:perf-per-labels}, this means that the addition of examples mainly benefits certain classes, and that these are well represented in the corpus.

\subsection{Label Definitions and Clarification Between Labels (RQ2)}

With globally far fewer tokens (1,063), our results indicated a higher performance when employing the label definitions (See~\textit{definition} in Table~\ref{tab:mod-perf}) than providing just the examples. 
This suggests that the model learns better from definitions than from examples because the classes are difficult to explain with examples in our case. This opens a big question about heuristics addressed to the process of selection of examples.
The impact of introducing clarifications about annotator errors and label ambiguities (\textit{definition+clarification}), or 4-shot examples (\textit{definition+examples}), 
into the model to the definition does not bring any global improvements. 
Some classes seem to benefit but to the detriment of others. 
As reported by~\cite{savelka2023gpt4}, the model does not appear to assimilate knowledge from annotators' mistakes and the inherent ambiguity, at least when they are presented as mere declarations.

\subsection{Textual Context Enrichment (RQ3)}

This experiment starts from the \textit{definition} configuration and studies the impact of adding textual context to the target sentence. 
By adding unlabelled sentences from the context (\textit{context$[-+]\backslash d$} in Table~\ref{tab:mod-perf}), we can see that the performance is deteriorating. 
However we see that any configuration with contextual sentences give better performance than zero-few shot examples, and that contextual sentences augmented with labels  (\textit{labelled\_c$[-+]\backslash d$})
outperform any of our experimented prompts.
A possible explanatory hypothesis may come from the degree of similarity shared by these various types of sentences with the target sentence \cite{liu-2022-good-examples,ye-2023-incontext-instruction}. Indeed few-shot (i.e. labelled examples that do not come from the document), unlabelled contextual sentences and labelled\footnote{On average, 70\% of the sentences that make up the 8 sentences preceding a sentence have the same label as that sentence. 80\% for 2 preceding sentences.} contextual sentences can be seen as three types of examples with increasing similarity and precision.
Notably, across all the contextual experiments, we observed that our results consistently improved when adding the following context compared to preceding context.

\subsection{Encouraging General or Specific Reasoning (RQ4)}

\begin{figure}[ht]
\centering
\begin{minipage}{.35\textwidth}
  \centering
  \includegraphics[width=1\textwidth]{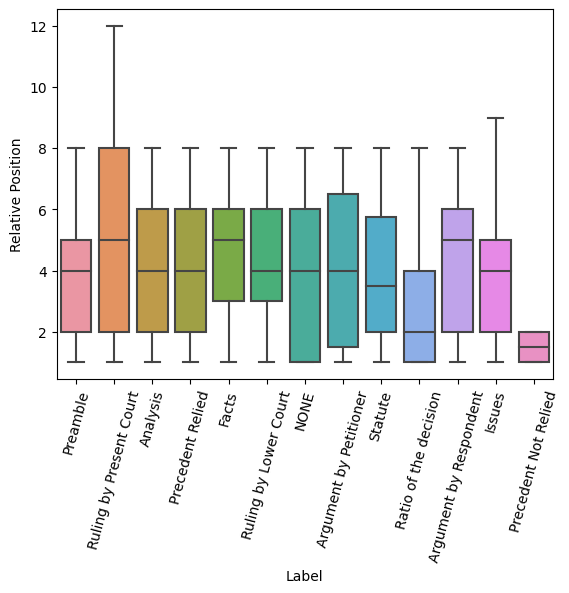}
  \captionof{figure}{Box Plot Illustrating the Distribution of Relation Positions by Labels}
  \label{fig:box-plot}
\end{minipage}%
\begin{minipage}{.69\textwidth}
  \centering
  \includegraphics[width=1\textwidth]{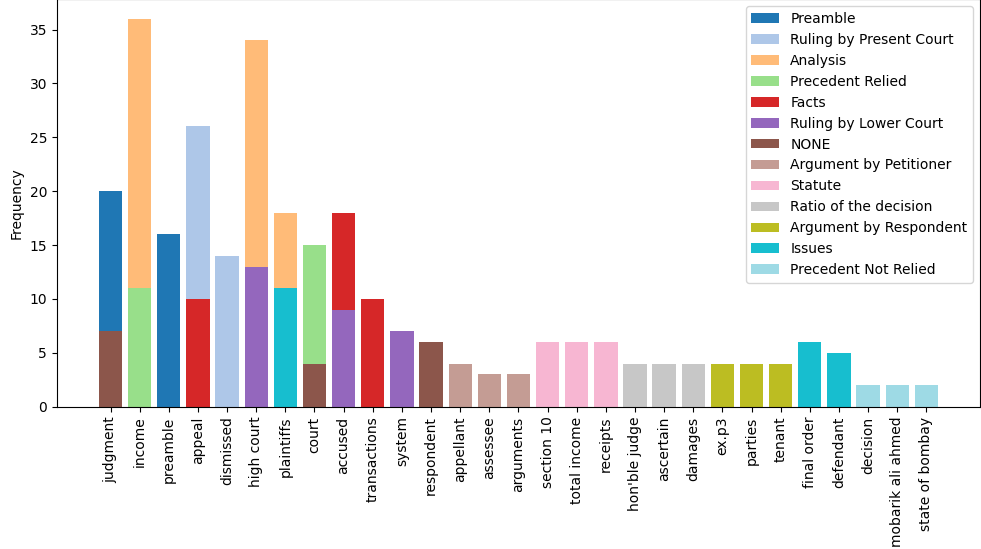}
  \captionof{figure}{Top 3 terms For each Rhetorical Roles}
  \label{fig:term-freq}
\end{minipage}
\end{figure}

Regarding the experimentation with a general reasoning instruction, 'Let's think step by step' (\textit{zero-shot-cot}), added to the \textit{definition} configuration, the model performs even worse than a zero-shot prompting.
These results confirm what was discussed in~\cite{savelka2023gpt4}, which indicated that GPT-3.5 struggles with correctly interpreting the annotation guidelines. 
When we targeted questions about the best prompt with context (\textit{labelled\_context+8}), we achieved higher performance compared to all the prompt strategies with an F1 score of 0.72 (\textit{cot-by-queries}). 
In the analysis of the targeted question about the relative position of the most influential sentence (Figure~\ref{fig:box-plot}), over 50\% of sentences in the RPC, FACTS, and ARG-RES roles are impacted by sentences located at least at position 5 within the context. 
However, the RATIO and PRE-NOT-RELIED roles have a lower median sensitivity, suggesting that they can be effectively recognized with shorter context sentences. Furthermore, as shown in Figure~\ref{fig:term-freq}, certain terms, such as "income," occurring in both the Analysis and Precedent roles, and "Plaintiffs," appearing in both Analysis and ISSUES roles, led to confusion, as discussed before in \cite{kalamkar2022corpus, malik2021semantic}. Additionally, terms like "preambule" and "dismissed" were found to be specialized for specific roles (PREAMBLE and RPC).

\subsection{Comparison with state-of-the-art}

The results we report from the state-of-the-art systems (See the bottom 4 rows of Table~\ref{tab:mod-perf}) concern systems which were fine-tuned with at least 25,800 pairs of examples. 
Through our experiments (in particular the labelled context ones) we show that a generative LLM, prompted in one stage, can outperform a supervised fine-tuned multi-class classifier based on the Transformer encoder model (\textit{BERT}). 
Although artificial, it opens the way to research.
However it seems difficult for such a generative system fed with classical prompts to beat a fine-tuned system with a context representation (i.e. \textit{BERT+local\_context}, \textit{SciBERT-HSLN} and \textit{AntContentTech}).

\section{Conclusion and Future work}

This study assessed the capabilities of GPT-3.5 in analyzing legal cases for the task of rhetorical roles prediction. 
We show that the number of examples, the definition of labels, the presentation of the textual context and specific questions about this context have a positive influence on the performance of the model.
In an artificial experiment, we observed that prompting with a few labelled examples from direct context can lead the model to a better performance than a supervised fined-tuned multi-class classifier based on the BERT encoder  (weighted F1 score of~$\approx$72\%). 
But there is still a gap to reach the performance of the best systems ~$\approx$86\%) in the LegalEval 2023 task which, on the other hand, require dedicated resources, architectures and training.

\section*{Acknowledgments}

This research was funded, in whole or in part, by l’Agence Nationale de la Recherche (ANR), project ANR-22-CE38-0004. 
\bibliographystyle{vancouver}
\bibliography{jurix2023}

\end{document}